\documentclass[10pt,twocolumn,letterpaper]{article}

\usepackage[pagenumbers]{iccv} 

\definecolor{iccvblue}{rgb}{0.21,0.49,0.74}
\usepackage[pagebackref,breaklinks,colorlinks,allcolors=iccvblue]{hyperref}

\usepackage{makecell}
\usepackage{graphicx}
\usepackage{amsmath, bm}
\usepackage{tabularx,booktabs}
\usepackage{adjustbox}
\usepackage{multirow}
\usepackage{graphicx,xcolor,colortbl} 
\definecolor{LightCyan}{rgb}{0.88,1,1}
\usepackage{array}
\newcolumntype{C}[1]{>{\centering\arraybackslash}p{#1}}
\newcolumntype{L}[1]{>{\arraybackslash}p{#1}}
\usepackage{subcaption}
\usepackage{indentfirst}
\usepackage{pifont}
\usepackage{lipsum}

\newcommand\blfootnote[1]{%
  \begingroup
  \renewcommand\thefootnote{}\footnote{#1}%
  \addtocounter{footnote}{-1}%
  \endgroup
}

\def\paperID{6019} 
\def\confName{ICCV}
\def\confYear{2025}

\title{DoubleDiffusion: Combining Heat Diffusion with Denoising Diffusion for Texture Generation on 3D Meshes}

\author{
Xuyang Wang$^1$ \quad 
Ziang Cheng$^2$ \quad 
Zhenyu Li$^3$ \quad
Jiayu Yang$^2$ \quad
Haorui Ji$^1$ \quad \\
Pan Ji$^2$ \quad
Mehrtash Harandi$^4$ \quad
Richard Hartley$^1$ \quad
Hongdong Li$^1$ 
\\[4pt]
$^1$~The Australian National University \\
$^2$~Tencent XR Lab \\
$^3$~King Abdullah University of Science and Technology\\
$^4$~Monash University
\vspace*{-10pt}
}

\begin{document}
\maketitle


\begin{abstract}
This paper addresses the problem of generating textures for 3D mesh assets. Existing approaches often rely on image diffusion models to generate multi-view image observations, which are then transformed onto the mesh surface to produce a single texture. However, due to the gap between multi-view images and 3D space, such process is susceptible to a range of issues such as geometric inconsistencies, visibility occlusion, and baking artifacts. To overcome this problem, we propose a novel approach that directly generates texture on 3D meshes. Our approach leverages heat dissipation diffusion, which serves as an efficient operator that propagates features on the geometric surface of a mesh, while remaining insensitive to the specific layout of the wireframe. By integrating this technique into a generative diffusion pipeline, we significantly improve the efficiency of texture generation compared to existing texture generation methods. We term our approach DoubleDiffusion, as it combines heat dissipation diffusion with denoising diffusion to enable native generative learning on 3D mesh surfaces.
\end{abstract}

\blfootnote{ \url{https://github.com/Wxyxixixi/DoubleDiffusion_3D_Mesh}}
    
\section{Introduction}
\label{sec:intro}

Texture generation task takes in an uncolored 3D object and output the texture color on the 3D object. The input object can be any of the 3D representations, such as Nerf, Gaussian Splatting, Point Clouds, Mesh, etc. While many methods target at Nerf~\cite{metzer2023latent, pooledreamfusion} and Gaussian Splatting~\cite{tang2024lgm} based texture generation, fewer methods have explored the texture generation on 3D mesh. Since meshes are widely used as a 3D representation in graphic design, generating texture directly on a given mesh has become an essential approach. 

Due to the lack of a unified 3D foundation model for meshes, some prior works~\cite{chen2023text2tex, yu2023texture} unwrap the surface into 2D texture maps and harness powerful image priors such as Stable Diffusion~\cite{rombach2021highresolution}. Although UV maps lie on a 2D plane, they are having geometric distortion and the texture cannot be generated directly via a 2D diffusion model. To bridge this gap, some methods takes the multi-view texture generation method as remedy, by using the 2D view images as the guidance for the objects' renders with the generated texture~\cite{chen2023text2tex,bensadoun2024meta, tang2024mvdiffusion, shimvdream}. Despite leveraging strong 2D priors, these methods require considerable computational resources to maintain multi-view consistency and often suffer from edge inconsistencies. Furthermore, these methods are not native mesh approach for texture generation.

\begin{figure}[t]
    \centering
    \includegraphics[width=\linewidth]{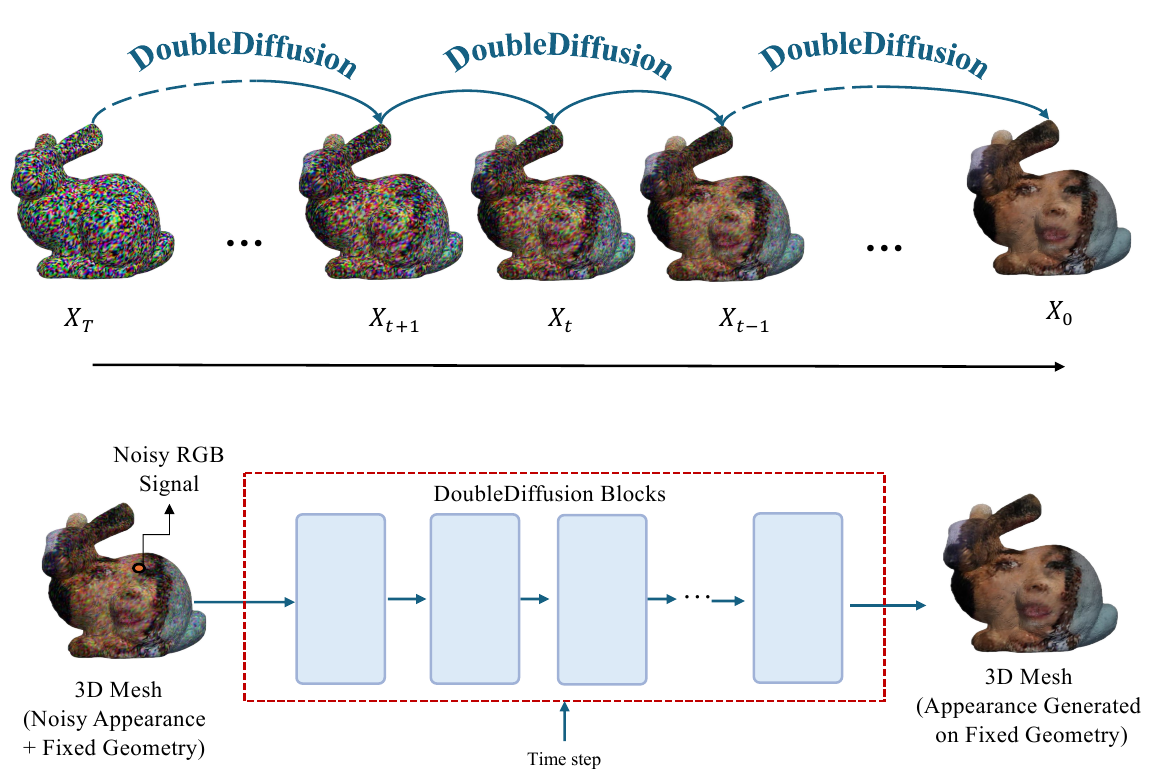}
    \caption{Overview of the DoubleDiffusion.}
    \label{fig:teaser}
\end{figure}

In our method, we frame texture generation on a 3D mesh as a signal generation problem on a curved surface. Unlike image generation on flat grids, the curvature of the geometric mesh plays a crucial role, making it impractical to directly apply a standard 2D convolutional operator. Although modern diffusion models have made significant progress in image domains -- thanks to convolutional or sequential networks operating on regular grids -- it remains challenging to transfer these convolutional kernels or auto-regressive modules directly to meshes. While some methods~\cite{elhagmanifold,foti2025uv} address this by applying positional encoding to sampled points to impose an order for signal learning, our approach instead leverages the native vertex-edge data structure.

Generating signals directly on the mesh, however, poses its own challenges. Mesh data structures are irregular, and each vertex’s neighborhood relationships lack a simple pattern, demanding a unified operator that respects the underlying geometry. We treat texture generation on the mesh as learning a color distribution anchored in the mesh’s geometry. Since the mesh is naturally a structured manifold where geodesic distances are defined by vertex connectivity, manifold operators are well-suited to capturing signal distributions on its surface. In mesh processing, the Laplacian-Beltrami (LB) operator~\cite{zhang2010spectral} serves as a second-order differential operator that measures how a scalar function varies with respect to the surface geometry. Its discrete form can process signal variation on the mesh, and Sorkine~\cite{sorkine2005laplacian} further shows that applying the discrete LB operator to the vertices corresponds to their mean curvature. Using the cotangent Laplacian, which encodes differential coordinates, effectively represents the local relationships between a vertex and its neighbors.

By viewing the texture signals on the manifold as a form of “heat energy,” we can apply the heat diffusion kernel to capture how signals evolve with respect to geometric neighbors. This concept parallels using convolutional or transformer layers in image-based diffusion models, except that our “noise predictor” in the denoising diffusion pipeline is designed for mesh geometry. Specifically, we embed a heat-diffusion-based network, DiffusionNet~\cite{sharp2022diffusionnet}, into a denoising diffusion framework to model RGB signal distributions directly on the mesh surface.

Albeit this straightforward design, direct operation on mesh proves highly efficient at learning RGB distributions while maintaining fast computation. We evaluate its performance on two types of generation tasks. First, we showcase how our method handles complex signal distributions on a single curved manifold, illustrating how geometry-aware diffusion enhances texture generation. Second, we test on the ShapeNet Core Dataset, which contains 50 categories, each with thousands of shapes, to assess the method’s capacity to learn textures across varying geometry and category-specific appearances. Notably, our pipeline processes batches of different shapes, learning the joint distribution of geometry and category-constrained texture.

In summary, our key contributions are:
\begin{itemize}
    \item We present DoubleDiffusion, the first framework for denoising diffusion probabilistic model that directly learns and generates signals on the 3D mesh surface, and achieves smoothness and view consistency along the mesh geometry. 
    \item We compare with the current SOTA manifold based generation method MDF~\cite{elhagmanifold} and achieves a significant 312.82\% improvement in coverage. Meanwhile, it can process large meshes of size \textit{e.g.,} $\sim$100k vertices in a single pass. It is also 8.1x faster to generate a sample on the manifold mesh. 
    \item Experiments on RGB distribution over a single manifold and on per-category shape-conditioned texture generation across multiple shapes demonstrate DoubleDiffusion's versatility. These results highlight the framework’s capability for geometry-adaptive signal generation on complex 3D surfaces.
\end{itemize}

\section{Related Work}
\label{sec:related_work}





\subsection{Texture Generation Directly on 3D Representations}
In 3D texture generation, challenges include the limited availability of high-quality 3D datasets and the high computational cost of training general diffusion models on mesh geometries. Some methods attempt to address these issues by projecting multiple views of the 3D object~\cite{zhang2024clay, bensadoun2024meta} and applying 2D image diffusion models~\cite{zhang2023adding, rombach2022high} to synthesize signals on the mesh. However, these approaches often result in inconsistency in overlapping regions. Another set of approaches generates content on alternative 3D representations, such as Gaussian Splatting~\cite{tang2025lgm}, Triplane~\cite{wang2024crm}, and Hash-grid~\cite{deng2024flashtex}, with subsequent mesh extraction, which frequently introduces artifacts and reduces quality~\cite{tang2025lgm, deng2024flashtex}. Differently, our work addresses these limitations by providing a framework for generating signals directly on 3D meshes in a way that respects the underlying geometry. This approach enables consistent, geometry-aware generation without the need for view-based projection or post-extraction process, thus preserving mesh integrity and fidelity.

\subsection{Differential Geometry and Mesh Processing}
In differential geometry, the surface of a 3D mesh is treated as a manifold, making learning over manifolds a longstanding topic of research in mesh processing. Taubin\etal~\cite{taubin1995signal} proposes using the Laplacian-Beltrami (LB) operator to define differential coordinates on meshes, noting that the integration of these coordinates over a small region approximates the mean curvature at a central point on the manifold. Later, Sörkine\etal~\cite{sorkine2005laplacian} demonstrates that a mesh’s geometry can be effectively captured using the discrete Laplacian, enabling detailed geometric analysis on non-Euclidean surfaces. \cite{zhang2010spectral} indicates that the LB operator offers a robust approach for learning signals over mesh geometry. Inspired from this, our method follows the geometry processing theory and make use of the mesh Laplacian operators to process the signals over the manifold. 

\subsection{Diffusion on Manifold}
Denoising Diffusion Probabilistic Models (DDPMs) are well-suited for learning on grid-like structured data, where neighborhood relationships are straightforwardly formulated with Euclidean distances. However, limited work has explored DDPMs within the context of manifold geometry. Manifold Diffusion Field (MDF)~\cite{elhagmanifold} and Diffusion Probabilistic Field (DPF)~\cite{zhuang2023diffusion} address signal generation over the manifold through a field representation with the coordinate-signal pairs. In particular, MDF replaces Cartesian coordinates adopted in DPF with Laplacian eigenvectors to represent manifold geometry, and uses Transformer as the denoising network. Unlike these field-based  approaches, our method directly processes signal distributions on manifold, integrating heat dissipation diffusion to capture local geometric dependencies and streamline the generative process. 

\subsection{Heat Diffusion}
The heat equation has been widely applied in message propagation networks, particularly within graph-based deep learning. A closely related work by Xu et al.~\cite{xu2020graph} leverages the heat kernel for semi-supervised learning on graphs, demonstrating that the Laplacian approximation over a heat kernel effectively filters out high-frequency components in node features, thereby enhancing signal smoothness. Similarly, Sharp\etal\cite{sharp2022diffusionnet} introduces a heat dissipation diffusion network specifically for mesh structures, providing an ideal tool for our approach. In this work, we integrate the heat diffusion network~\cite{sharp2022diffusionnet} with a probabilistic diffusion generative model. Here, heat diffusion serves to suppress highly varying components in the noisy feature space, acting as an effective denoiser for the denoising diffusion probabilistic model on meshes.


\section{Preliminary}

\subsection{Heat Diffusion}
Heat diffusion has long been used in describing how an initial scalar heat distribution $\bm h$ diffuses or ``spread out'' over time `$s$'. To distinguish the heat-diffusion time 
`$s$' from the denoising diffusion timestep $t$, we use $s$ to represent the duration of heat diffusion. In differential geometry, heat diffusion on a mesh surface follows:{\begin{equation}
    \frac{\partial\bm h(s)}{\partial s} = \Delta \bm h(s),
    \label{eq:heat_diffusion}
\end{equation}}
where $\bm h(s)$ denotes the heat or feature energy across the surface, and $\Delta$ is the Laplacian Beltrami operator that characterizes how temperature at each point diverges from its surroundings. The heat energy spreading speed depends on the design of the Laplacian matrix and the heat function. In our method, we used the DiffusionNet~\cite{sharp2022diffusionnet}, which supports regions of greater curvature or density experience more extensive feature-energy exchange during the diffusion process.

\begin{figure}[t]
  \centering
  \begin{subfigure}{0.33\linewidth}
    \includegraphics[width=0.8\linewidth]{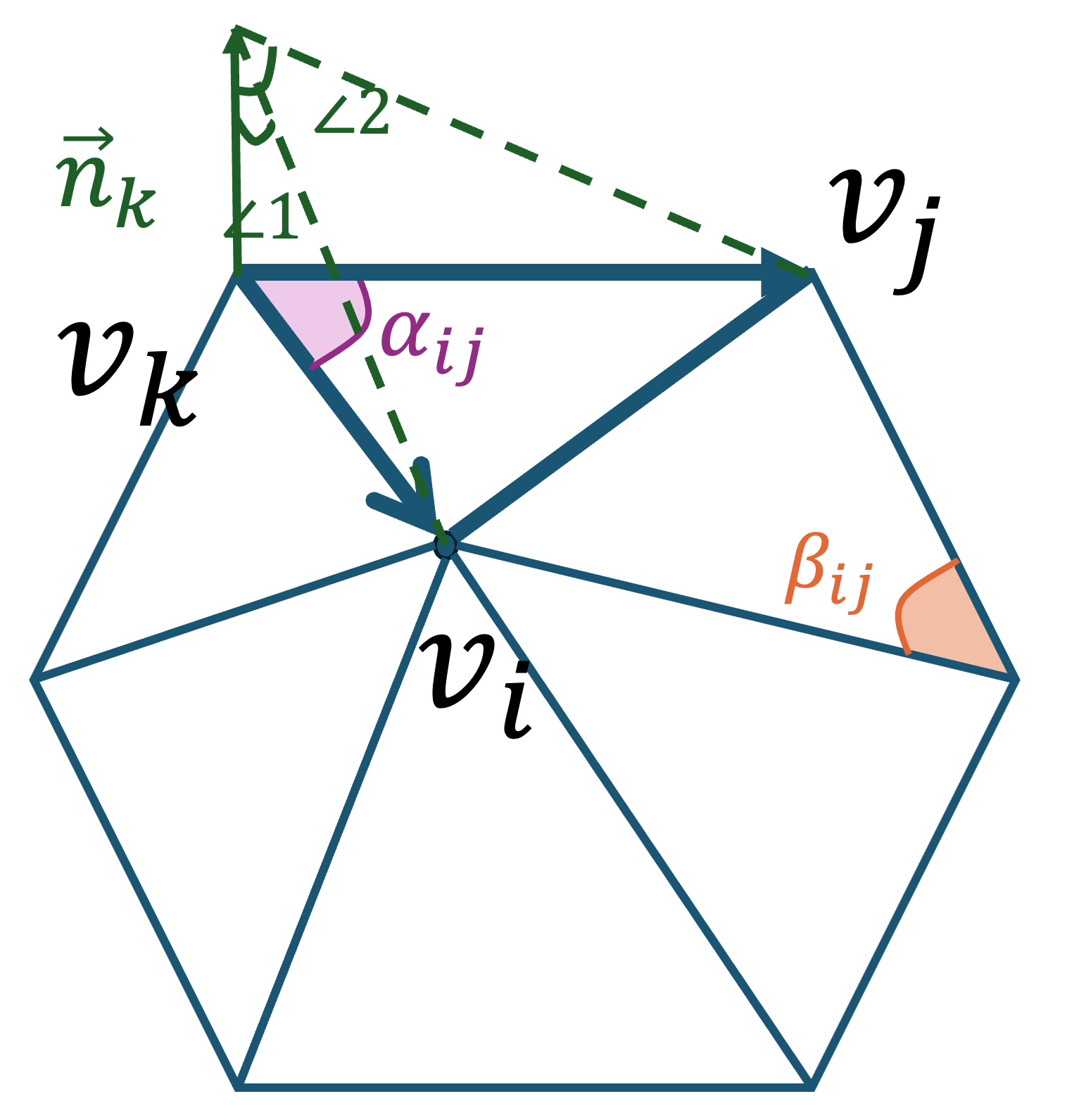}
    \caption{Cotangent weights calculated on vertex $v_i$ w.r.t to edge (i,j)}
    \label{fig:cotan_weight_fig}
  \end{subfigure}
  \hspace{2mm}
  \begin{subfigure}{0.62\linewidth}
    \includegraphics[width=0.8\linewidth]{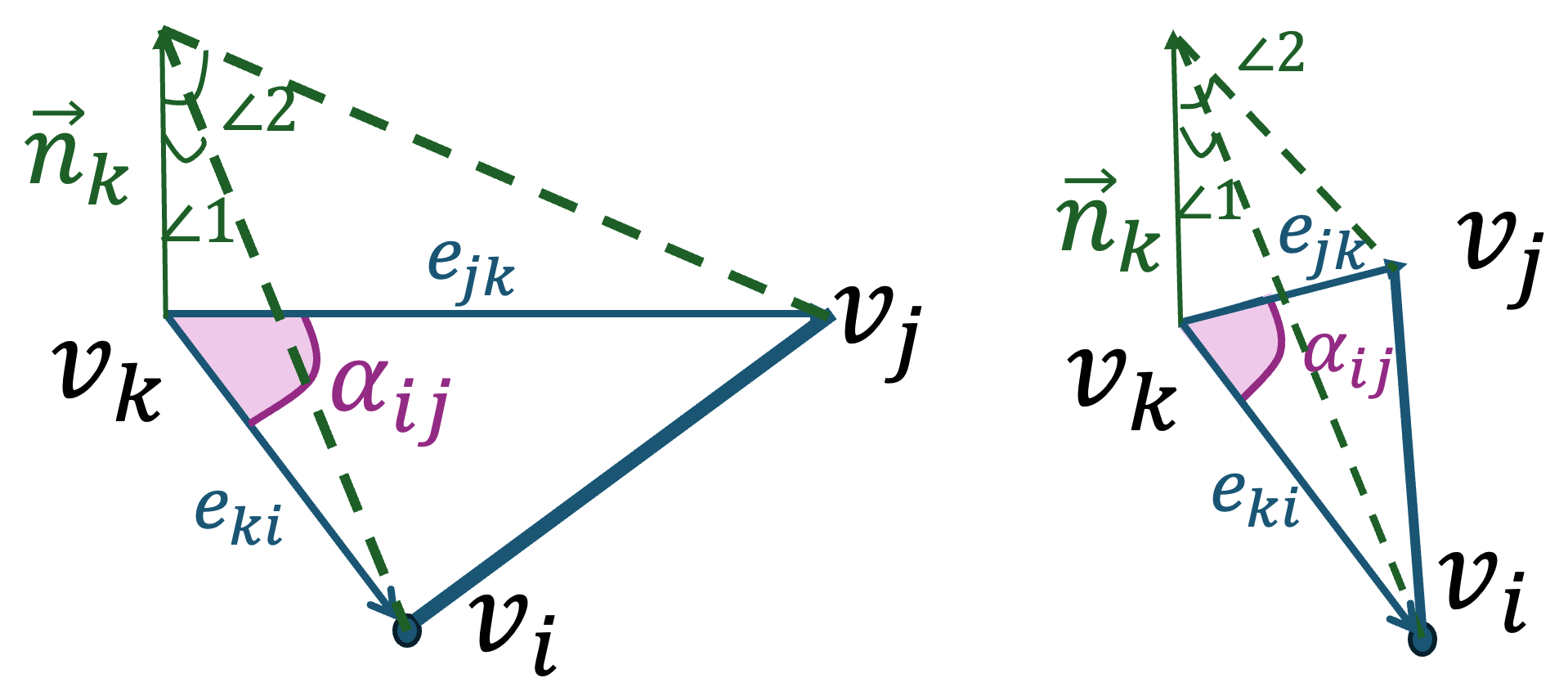}
    \caption{Cotangent of $\angle \alpha_{ij}$ is proportional to the area of triangle $F_{ijk}$, larger area in the left triangle will also result in a larger $\cot \alpha_{ij}$ comparing to the triangle in the right.}
    \label{fig:cotan_fig}
  \end{subfigure}
  \caption{Illustration of the cotangent weighted differential coordinates on the manifold with respect to manifold mesh.}
  \label{fig:cotan_weight}
\end{figure}

\subsection{Mesh Laplacian}
\label{sec:cotangent}
In our method, we adopt the cotangent-weight Laplacian matrix~\cite{meyer2003discrete} as the Laplacian Beltrami Operator in Eq.~\ref{eq:heat_diffusion}. A typical mesh consists of vertices, edges, and faces. In graph learning, the graph Laplacian has a long history of usage, focusing primarily on node connectivity. By contrast, the cotangent-weight Laplacian matrix has been widely used as a geometric mesh Laplacian, where the edge weights depend not only on edge lengths but also on the areas enclosed by triangular faces. Consequently, it is loosely related to the Gaussian curvature at the central vertex $v_i$~\cite{meyer2003discrete}.

Applying cotangent-weight Laplacian Beltrami Operator gives:
\begin{equation}
    [\mathbf{L} \bm h]_i = \frac{1}{\Omega_i} \sum_{j\in N(i)} \frac{1}{2}(\cot \alpha_ij + \cot \beta_ij)(h_i - h_j) 
\label{eq:laplace}
\end{equation}

where $\mathbf{L}$ represent the Laplacian Beltrami Operator, the $\Delta$ in Eq.~\ref{eq:heat_diffusion}. $\bm h_i$ represent the heat energy that being received for the vertices i. $\Omega_i$ is the total area of the triangle that contains vertices i, $N(i)$ are the vertices that is neighbour to the vertices i, and $\alpha_{ij}$, $\beta_{ij}$ represents the angles that opposite to the edge $e_{ij}$.
\section{Method}
\subsection{Overview of DoubleDiffusion}
\label{sec:overview}


\begin{figure*}[t]
\setlength\tabcolsep{1pt}
\centering
\small
    \begin{tabular}{@{}L{4cm}L{4cm}L{4cm}L{4cm}@{}}
    \multicolumn{3}{c}{\includegraphics[width=\linewidth]{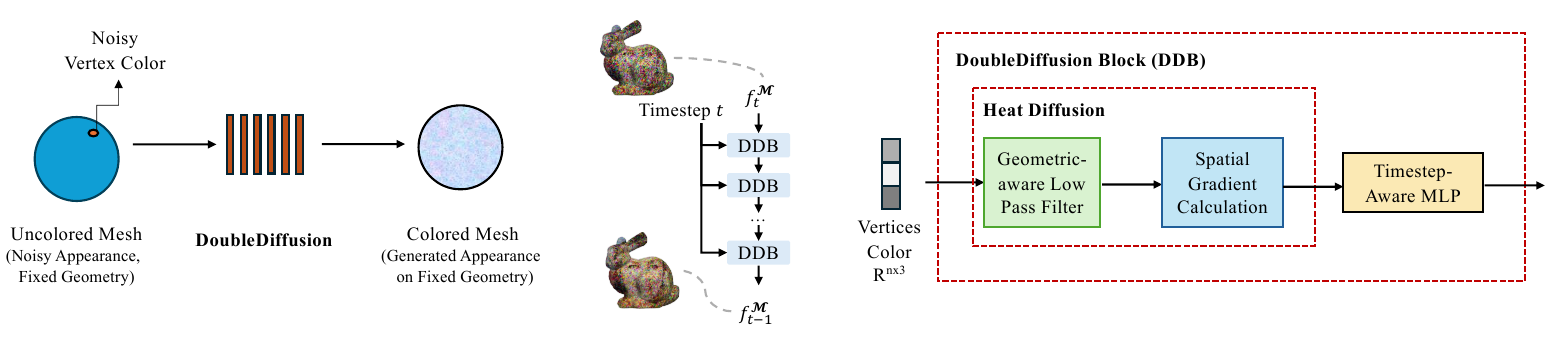}} \vspace{-0.2cm}\\
    ~~~~~~~~~~~~~~~~~~~~~~~~~~~~~~~~~~~~(a) & ~~~~~~~~~~~~~~~~~~~~~~~~~~~~~~~~~~~~~~~~~~~~(b) & ~~~~~~~~~~~~~~~~~~~~~~~~~~~~~~~~~~~~~~~~~~~~~~~~~~~~~~~~~~~~~~~~~~~~~~~~~~~~~~~~~~~~~~~~~~(c) \\
    \end{tabular}
    \vspace{-0.3cm}
    \caption{\textbf{Method Overview.} Our (a) DoubleDiffusion Network contains consecutive (b) DoubleDiffusion Blocks (DDB). Each (c) DDB consists of the  Heat Diffusion module aggregating vertice features over the spatial domain on mesh surfaces and the Timestep-Aware MLP that injects timestep embedding in a residual manner.}
    \label{fig:method}
\end{figure*}


The DoubleDiffusion idea is simple and straightforward. First of all, we have an uncolored mesh $\mathcal{M} = (\mathbf{V}, \mathbf{F})$ with $n$ number of vertices $ \{\bm{v}_0, \bm{v_1}, \cdots, \bm{v_n}\} \in \mathbf{V} $. And we describe the vertices color or the vertices features as the scalar function $\bm f$ resides on the manifold mesh, $ \bm{f} = (\bm{f}^{v_0}, \bm{f}^{v_1}, \cdots, \bm{f}^{v_n}) $, where $ \bm{f}^{v_1} $ refers to the scalar vector on vertex $ \bm{v}_i $. Here, the $\bm f$ is having the similar meaning to the $\bm h$ in Eq.~\ref{eq:heat_diffusion}, in our task, $\bm f$ are initially sampled from the Gaussian Noise $\bm f_T \sim \mathcal{N}(0,1)$. These noisy vertex color will go through the DoubleDiffusion Blocks (DDB) iteratively to be denoised and generate the texture color as shown in Fig.~\ref{fig:method}a and Fig.~\ref{fig:method}b. 

Each of the DoubleDiffusion Blocks contains a Heat Diffusion and a Timestep-Aware per Vertex MLP as shown in Fig~\ref{fig:method}. The design of the DDB adapts from DiffusionNet~\cite{sharp2022diffusionnet}, which has been demonstrated as a powerful surface processing network recently~\cite{harrison2024improving}.

\subsection{Heat Diffusion Module}
\label{sec:heat_diffusion_module}
The Heat Diffusion Module contains two main parts. The first part is a geometric-aware low pass filter, which help to spread out the feature to the vertices resides on the geometric mesh. The second part is a spacial gradient calculation, which accelerate the heat diffusion on the surface with respect to the surface gradients.

\subsubsection{Geometric-aware Low Pass Filter}
\begin{figure*}[t]
    \centering
    \includegraphics[width=\linewidth]{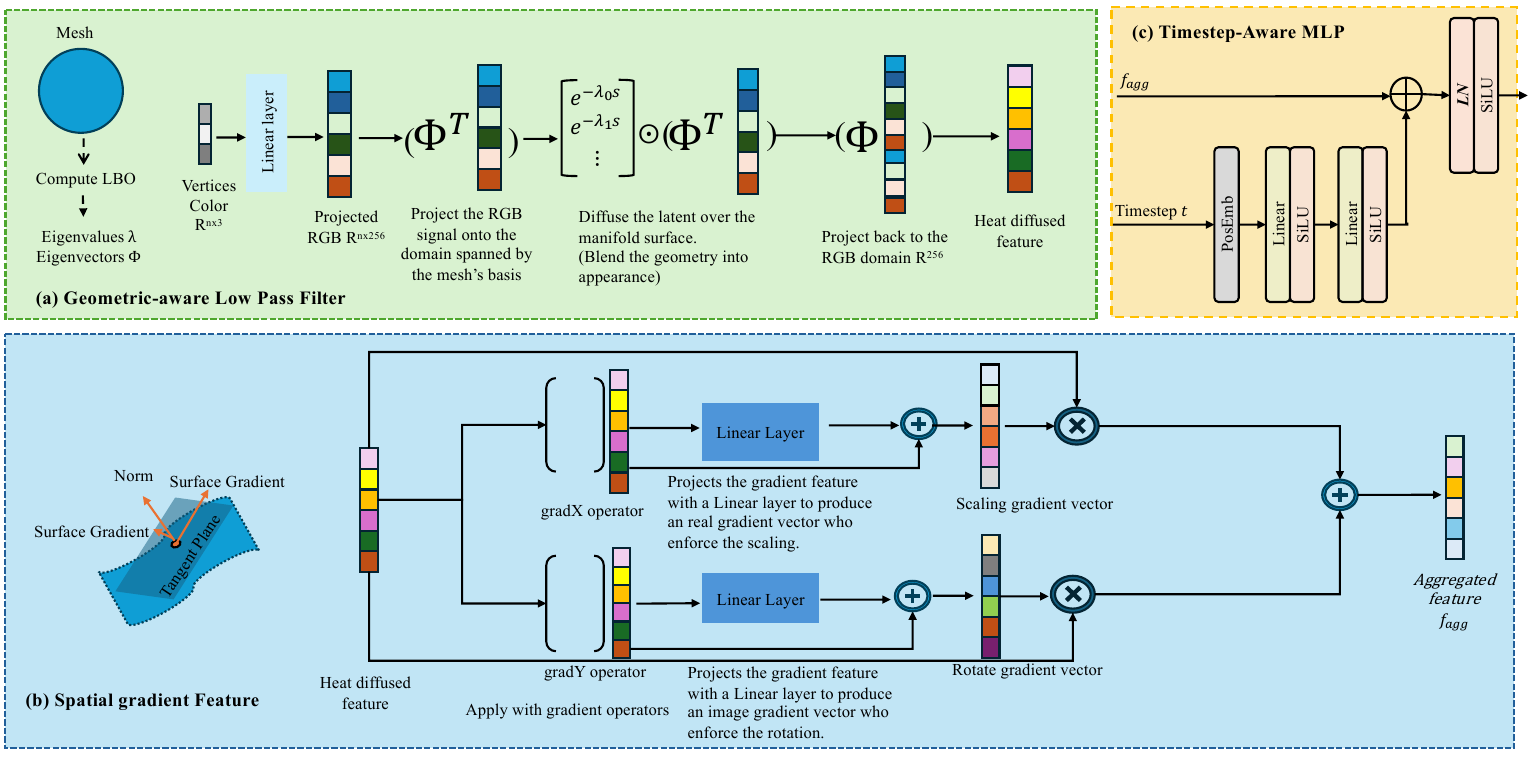}
    \caption{Overview of the DoubleDiffusion Block, containing (a) Geometric-aware Low Pass Filter; (b) Spacial gradient Feature and (c) Timestep-aware MLP.}
    \label{fig:ddb}
\end{figure*}

The geometric-aware low pass filter as shown in Fig~\ref{fig:ddb}~(green) is a Laplacian approximation of the heat diffusion. Solving the heat equation presents an eigenvalue problem, where we can use the eigenvalue decomposition to approximate the heat diffusion. To apply the heat diffusion, we need heat kernel, which is defined as
\begin{equation}
    h(\lambda_i) = e^{-s\lambda_i}
    \label{eq:heat_kernal}
\end{equation}
where $s\geq 0$ is a scaling hyper-parameter. The heat kernel can define the convolutional kernel on the surface which is $e^{-sL}$. To compute the heat kernel, we can use the eigen-decomposition to approximate this heat kernel, yielding:

\begin{equation}
    L = \Phi \Lambda \Phi^T,
\label{eq:heat_approximation}
\end{equation}
where $\Phi $ contains the eigenvectors of $L$, and $ \Lambda $ is a diagonal matrix with eigenvalues $\lambda_i$. 

The heat diffusion on the vertex color is then approximated as:
\begin{equation}
    \mathbf{H}(s) \bm{f} = \Phi e^{-s \Lambda} \Phi^T \bm{f} = \sum_{i=1}^{k} e^{-s \lambda_i} \langle \bm{f}, \phi_i \rangle \phi_i.
\label{eq:laplacian_heat}
\end{equation}

In this spectral representation, the heat diffusion kernel $e^{-s \Lambda}$ has diagonal elements $e^{-s \lambda_i}$ that control the rate of decay for each frequency component. This approximation also brings effect to fast compute to the heat diffusion Eq.~\ref{eq:heat_diffusion} on large meshes. Meanwhile, Laplacian operators has been recognized as a spectral filer~\cite{taubin1995signal,levy2006laplace}. Choosing the eigenvector of the mesh Laplacian $\mathbf{L}$ with $k$ smallest eigenvalues, Laplacian operator selectively attenuates high-frequency components (sharp changes or noise) while preserving low-frequency components (smooth, global variations). Thus, the Laplacian-approximated heat diffusion functions by dissipating high-frequency variations more quickly than low-frequency components. This behavior effectively smooths the signal, by discounting high-frequency basic filters via heat kernels.

Furthermore, we followed the method introduce in \cite{sharp2022diffusionnet}, to apply the heat kernel with respect to the feature channel:
\begin{equation}
    \mathbf{H}(s) \bm{f} = \Phi \begin{bmatrix} e^{-\lambda_0 s} \\ e^{-\lambda_1 s} \\ \vdots \end{bmatrix} \odot (\Phi^T \bm f),
\end{equation}


where the $\Phi$ only contains the $k$ smallest eigenvector of the $L^{MC}$ and the eigenvalues are in small to large order $\lambda_0 < \lambda_1 < \cdots < \lambda_k$. The symbol $\odot$ represents the element-wise multiplication. In this process, the $k$-eigenvectors first project the feature signals into the spectral domain, where the high-frequency components are suppressed by the frequency-modulated heat kernel, while low-frequency components are enhanced. This approach achieves channel-wise feature aggregation across vertices in the spectral domain, effectively gathering features within the heat diffusion range defined by the learnable $s$ (Fig.~\ref{fig:ddb}). 


\subsubsection{Spatial Gradient Calculations}
Following the DiffusionNet, the diffused feature for each vertex will then be multiplied by the gradient operators.  The structure of this module is shown in Fig.~\ref{fig:ddb}. The calculation of the gradient operators gradX and gradY matrix is calculated with respect to the vertex normals, and the details could be found in ~\cite{sharp2022diffusionnet}.

\subsection{Denoising Process with Heat Diffusion.} 
The DoubleDiffusion generate samples by starting with a noise distribution $ \bm{f}_T^{v_i} \sim \mathcal{N}(0,1) $, the initial noise is independently sampled at each vertex. And the linear noise scheduler~\cite{ho2020denoising} is used to prepare the noisy data. Then the DoubleDiffusion is served as the noise predictor $ \bm{\epsilon}_\theta(\bm{f}_t, t)$ and being trained to progressively removes noise through each step $ p_\theta(\bm{f}_{t-1} | \bm{f}_t)$.
\begin{equation}
p_\theta(\bm{f}_{t-1} | \bm{f}_t) = \mathcal{N}(\bm{f}_{t-1}; \mu_\theta(\bm{f}_t, t), \sigma_t^2 \, \mathbf{I}),\\
\end{equation}
\begin{equation}
\mu_\theta(\bm{f}_t, t) = \frac{1}{\sqrt{1 - \beta_t}} \left( \bm{f}_t - \frac{\beta_t}{\sqrt{1 - \alpha_t}} \bm{\epsilon}_\theta(\bm{f}_t, t) \right)
\end{equation}

Following previous work~\cite{rombach2022sd}, we incorporate the time step $t$ as a time embedding within each consecutive block using a per-vertex MLP. Given that heat diffusion is conducted on a per-channel basis, employing group normalization~\cite{wu2018group} directly disrupts the preservation of spatial information, resulting in non-convergence of the model. Consequently, we empirically adopt layer normalization~\cite{ba2016layer} to enhance the training process as shown in Fig~\ref{fig:ddb}(c).

\textbf{Loss Function.} We train by minimizing the MSE loss between the true noise $ \bm{\epsilon} $ and predicted noise $ \bm{\epsilon}_\theta(\bm{f}_t, t) $:

\begin{equation}
\mathcal{L}_{\text{DDPM}} = \mathbb{E}_{\bm{f}_0, \bm{\epsilon}, t} \left[ \left\| \bm{\epsilon} - \bm{\epsilon}_\theta(\sqrt{\alpha_t} \, \bm{f}_0 + \sqrt{1 - \alpha_t} \, \bm{\epsilon}, t) \right\|^2 \right],
\end{equation}
where $ \bm{\epsilon} \sim \mathcal{N}(0, \mathbf{I}) $ and $ t $ is sampled from $ \{1, \dots, T\} $.

\section{Experiments}
In this section, we evaluate our proposed framework, \textbf{DoubleDiffusion}, for learning signal distributions on 3D mesh manifolds. We conduct three main experiments: (1) learning RGB color distributions on one prescribed manifold Stanford bunny~(Sec.~\ref{exp:celea_bunny}), (2) learning the texture distribution over a set of prescribe manifold and generates diverse textures conditioned on each shape’s geometry~(Sec.~\ref{exp:shapenetcore}). Both tasks are designed to test out the direct generative ability on the manifold as well as the potentials of the proposed method.

\label{sec:exp}
\subsection{Distribution Generation on Manifold}
\label{exp:celea_bunny}
In this experiment, we evaluate the ability of our framework to learn complex RGB distributions on a 3D curved surfaces. The learning objective is to model the probability distribution $ p_\theta(f^{v_i}_{t-1} | f^{v_i}_{t}) $, where each $ v_i \in \mathcal{M} $, with $ \mathcal{M} $ representing the Stanford bunny manifold~\cite{turk1994stanfordbunny}. Here, $ f^{v_i}_{t-1} $ is the vertex feature at denoising step $ t-1 $, with the initial feature $ f_{t=0}^{v_i} $ being a 3-dimensional RGB value at each vertex acquired from Celeba-HQ human face dataset~\cite{karras2017progressive}. This setup allows us to assess the framework’s effectiveness in capturing a meaningful distribution over the manifold.

\begin{figure}[t]
  \centering
    \includegraphics[width=\linewidth]{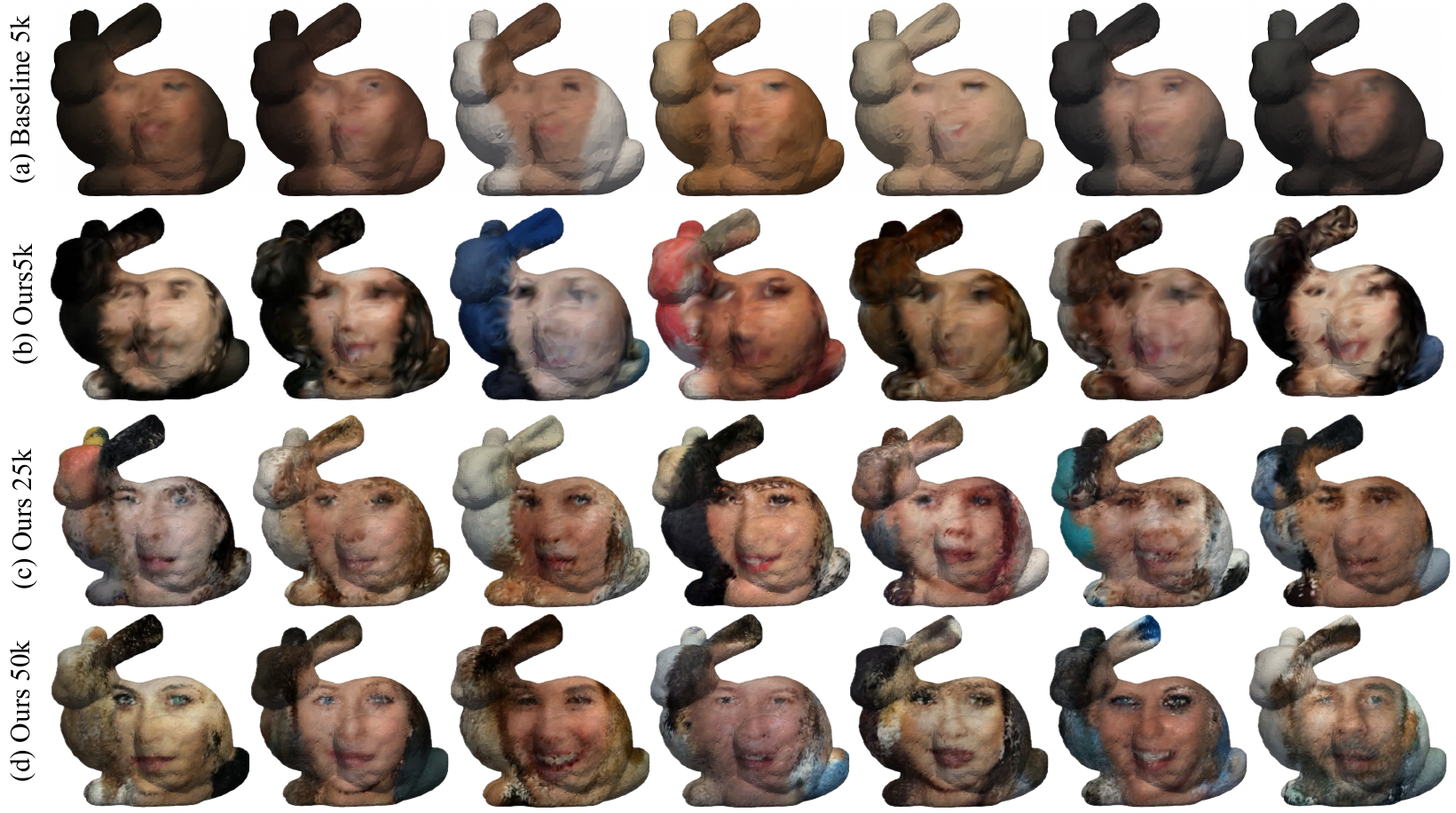}
    \caption{Qualitative comparison between the (a) MDF (baseline) and (b) DoubleDiffusion (ours) on the Stanford Bunny with $\sim$5k number of vertices. Our method can efficiently handle the mesh with large
number of vertices manifold mesh. Here we show our result with (c) $\sim$25k vertices, and (d) $\sim$52k vertices. The meshes with higher vertices are watertight remeshed with the off-the-shelf watertight manifold tool ManifoldPlus~\cite{huang2020manifoldplus}.}
  \label{fig:quali_4k}
\end{figure}

\begin{table}[t]
  \centering
  \begin{tabular}{@{}L{1.2cm}C{1.3cm}C{1.3cm}C{1.3cm}C{1.3cm}@{}}
    \toprule
    Method & MMD $\downarrow$ & COV $\uparrow$ &\#param.$\downarrow$ & $t$ (s) $\downarrow$ \\
    \midrule
    MDF  & 0.259 & 7.75  & 11.4M & 1.14\\
    Ours & 0.284 & 32.01 & 2.4M & 0.14\\
    \bottomrule
  \end{tabular}
  \caption{Quantity comparison between MDF (baseline) and DoubleDiffusion (ours) with CelebA-HQ on Stanford Bunny (number of vertices $\sim$5k). With comparable MMD, our method surpasses MDF in terms of COV with a large margin.Meanwhile, our model is 4.7x smaller in size than MDF and 8.1x faster for the inference, with only 0.14 second per denoising step.}
  \label{tab:quanti}
\end{table}

\subsubsection{Implementation}
\textbf{Dataset Preparation.} As there's not an available datasets with a large number of texture for one manifold, we follow the baseline method MDF to prepare the manifold value data source from the CelebaA-HQ image, as the CelebA-HQ~\cite{karras2017progressive} dataset has complex and diverse RGB distributions. Initial RGB values $ f_{t=0}^{v_i} $ are mapped onto the Stanford bunny manifold using an off-the-shelf texture mapping tool, PyVista~\cite{sullivan2019pyvista}, which provides a texture coordinate to the stanford bunny, and we then assigned the per-vertex color values from the image data in Celeba-HQ. Along with the manifold-color pair, we precompute the cotangent-weighted Laplacian matrix for the mesh, along with its eigenvalues and eigenvectors, to facilitate heat diffusion. The CelebA-HQ dataset consists of 30,000 images. 
We use the official split file with 24,183 images as the training set and 2,824 images as the test set.

\textbf{Training Details.} During training, the network predicts the denoising of RGB values across the manifold. The prediction network utilizes heat diffusion adapted from DiffusionNet~\cite{sharp2022diffusionnet} as a feature processing mechanism. We employ the EDM~\cite{karras2022elucidating} framework as the denoising probabilistic model for faster inference with 18 timesteps. All experiments are conducted on 4 NVIDIA A100s, with 96 training epochs, learning rates at 3e-2, batch size of 8, 128 eigenvector to represent the approximation of the heat dissipation diffusion and 8 DoubleDiffusion blocks. The inference time is evaluated on the NVIDIA A100 GPU.

\subsubsection{Evaluation}
\textbf{Baseline and Metrics.} Due to the absence of official source code for MDF~\cite{elhagmanifold}, we meticulously reproduce the baseline MDF method based on the detailed descriptions provided in their paper and its appendix. Specifically, we employed the Transf. Enc-Dec architecture as outlined in the appendix of MDF. We evaluated both MDF and our approach using a manifold-processed and simplified Stanford Bunny model, consisting of approximately 5k vertices~\cite{huang2020manifoldplus}. This vertex count is significantly lower than the 50k+ vertices our model is capable of handling. The limitation is due to the computational constraints imposed by MDF’s global attention mechanism, which aggregates features across vertices and escalates computational complexity. For our evaluations, we adopted the metrics proposed by Achlioptas \etal~\cite{achlioptas2018learning}: Coverage (COV) and Minimum Matching Distance (MMD). These metrics assess the alignment between the test set and the generated samples, and indicate the accuracy with which the test set is represented within the generated samples, respectively. We followed the implementation from the official code\footnote{ \url{https://github.com/optas/latent_3d_points/blob/master/notebooks/compute_evaluation_metrics.ipynb}}~\cite{achlioptas2018learning} to use the ground-truth and generated samples as reference and sample points, respectively, a practical contrary to that used by MDF. In particular, at equivalent levels of MMD a higher COV is desired~\cite{achlioptas2018learning}, and vice-versa. To have a fair comparison, we generate equal number of data samples as the test set to compare the generated distribution with the test distribution.

\begin{figure*}[t]
\setlength\tabcolsep{1pt}
\centering
\small
    \begin{tabular}{@{}L{4cm}L{4cm}L{4cm}L{4cm}@{}}
    \multicolumn{2}{c}{\includegraphics[width=0.85\linewidth]{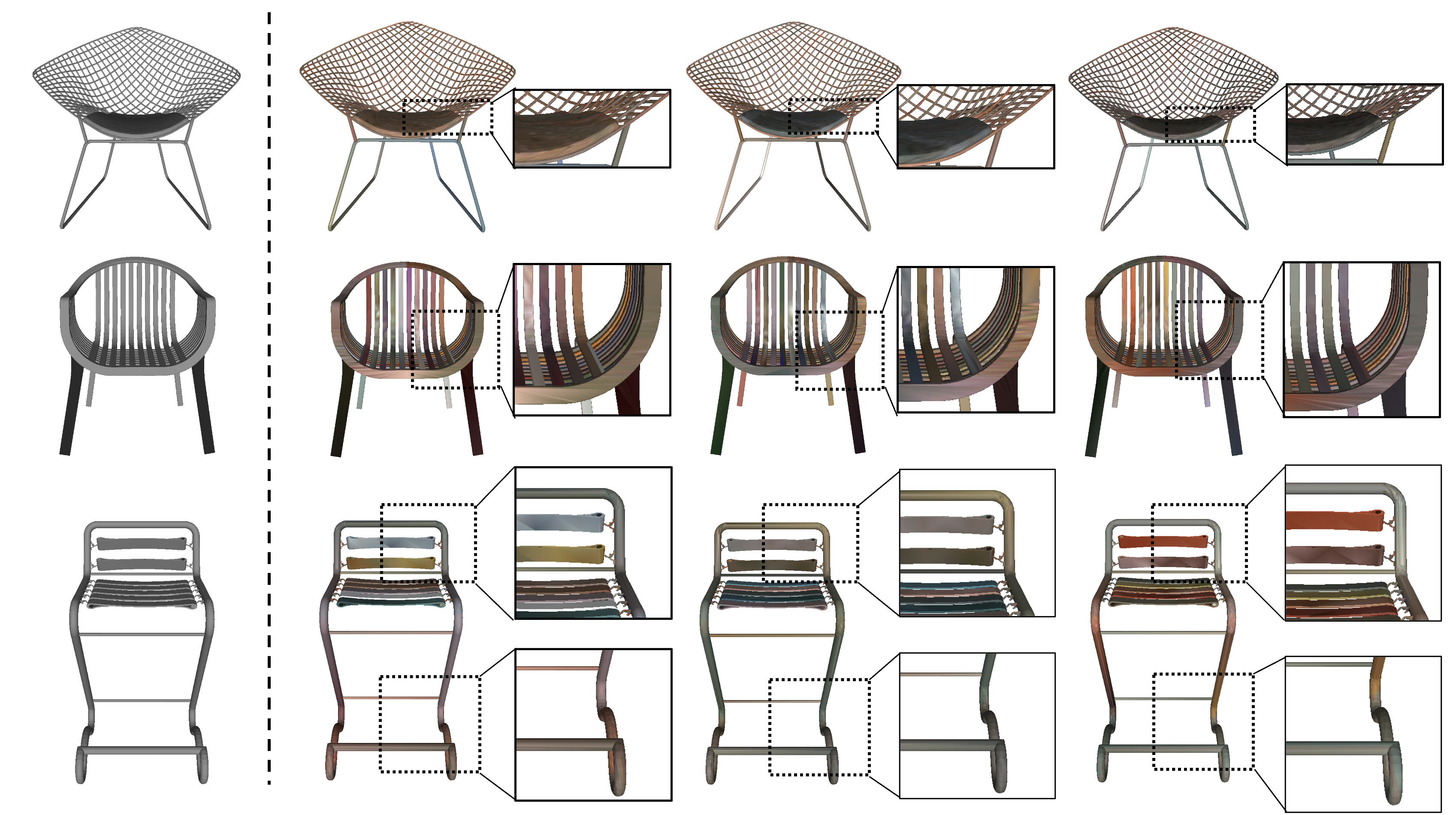}} \vspace{-0.2cm}\\
    ~~~~~~~~~~~~~~~~~~~(a) & ~~~~~~~~~~~~~~~~~~~~~~~~~~~~~~~~~~~~~~~~~~~~~~~~~~~~~~~~~~~~~~~~~~~~~~~~~~~~~~~~~~~~(b) \\
    \end{tabular}
    \vspace{-0.2cm}
  \centering
    \caption{Texture learning across multiple chair shapes in the \textbf{ShapeNetCore v2} ~\cite{chang2015shapenet} dataset, with (a) the uncolored chair meshes and, (b) the textured chair meshes sampled from DoubleDiffusion.} 
  \label{fig:shapenet_tex}
\end{figure*}

\textbf{Comparison.} We first present a quantitative comparison in Tab.~\ref{tab:quanti}. Although our method posts comparable Minimum Matching Distance (MMD) scores, it achieves a significant 312.82\% improvement in Coverage (COV), demonstrating superior alignment between the test-set and generated fields. Additionally, our model is 4.7 times smaller and 8.1 times faster than MDF, highlighting our efficient feature learning that leverages both heat diffusion and denoising diffusion processes. Fig.~\ref{fig:quali_4k} showcases the qualitative comparison results. Our approach produces textures with enhanced details and more distinct facial features. In contrast, the textures generated by MDF appear excessively smoothed, resulting in a loss of fine details and a uniformly bland, less expressive surface appearance, which we refer to as `average faces'. This phenomenon aligns with the quantitative findings, where MDF tends to score better on MMD due to its tendency to predict average-like features thus fails to cover the test-set distribution. This discrepancy can be attributed to MDF's reliance on Laplacian eigenmaps for spectral positional encodings and its formulation of texture generation as a field. During training, MDF must sample a limited number of points to stay within memory constraints, which may prevent it from fully capturing the underlying geometry. In contrast, our method facilitates local feature aggregation more attuned to the mesh's shape, resulting in textures that more accurately reflect the surface's curved structure.

\subsubsection{Apply to Larger Meshes}



We scaled our experiments to handle larger mesh with $24,992$ ($\sim$25k) and $52,299$ ($\sim$52k) vertices. Fig.~\ref{fig:quali_4k} demonstrates our method can successfully generate comparatively high-quality textures on these higher-resolution meshes. With the support of more vertices, the generated samples capture better details with less noise. This scalability sheds lights on generative learning over large mesh with complex manifold surfaces, making DoubleDiffusion suitable for applications requiring detailed texture generation over the large number of vertices' meshes, with a fast speed.

Note that it is \textit{\textbf{infeasible}} for MDF to handle such high-resolution meshes. Facing the limitation of scalability, MDF can be only trained with a maximum of approximately 5,000 vertices on our default GPU. It is due to the adoption of the global attention mechanism to aggregate features at the vertices~\cite{mitchel2024single}, which has quadratic time and memory complexity. While adopting the advanced framework like PerceiverIO~\cite{jaegle2021perceiver} can handle this issue, the performance decreases with the reduce of the number of content points~\cite{elhagmanifold,zhuang2023diffusion}. Moreover, during training, MDF randomly samples points on the mesh surface to facilitate training, which can also hinder the modal to learn the underlying complete mesh geometry. In contrast, our method is specifically designed to learn signal distributions directly on the 3D manifolds by integrating heat dissipation diffusion with a probabilistic model. This combination allows to make a full use of the advantage of heat diffusion to effectively learn on complete mesh shape, and the ability of the denoising diffusion to generate high-quality signals.


\subsection{Per-category Texture on ShapeNetCore}
\label{exp:shapenetcore}
\begin{figure}[t]
  \centering
    \includegraphics[width=\linewidth]{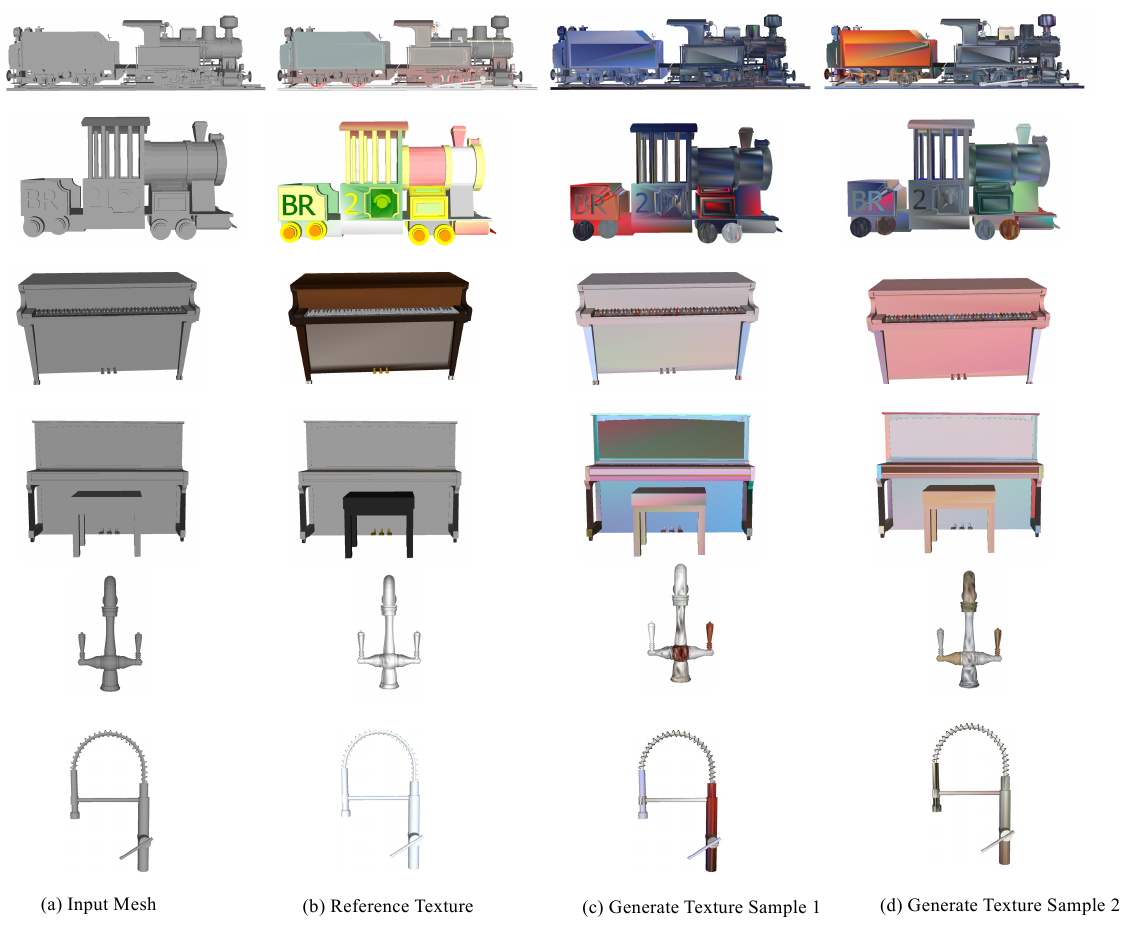}
    \caption{Generated results on different categories. Line 1-2 is the result on the train category, line 3-4 is the generated result for piano category, line 5-6 is the texture generation on the faucet category.} 
  \label{fig:multi-cate}
\end{figure}

In this experiment, we extend DoubleDiffusion to learn over multiple manifold shapes. This setup models the distribution $ p_\theta(f^{\mathcal{M}_{k, v_i}}_{t-1} | f^{\mathcal{M}_{k, v_i}}_{t}) $, where $ \mathcal{M}_{k, v_i} $ denotes vertex $ v_i $ on manifold $ \mathcal{M}_k $, with $ \mathcal{M}_k \in \{ \mathcal{M}_m \} $ representing a set of manifold shapes in the training data. This setup enables DoubleDiffusion to generate textures that are adapted to the structure of each individual mesh in the category, showcasing its capability to handle per-category texture generation across different manifolds.

\subsubsection{Dataset and Implementation Details}
We use the ShapeNetCore dataset~\cite{chang2015shapenet} to perform a per-category texture generation. ShapeNetCore is a 3D objects dataset with 55 different categories, such as chairs, caps, cars, etc. Each shape object in the dataset have one corresponding texture. We use the area interpolation to acquire the texture color on the vertices of the shapes, as the texture atlas are defined in per-face manner. As a demonstration, we trained on the ``chair" category, which includes 2,412 training samples and 311 test samples. We use the same network setting with the single mesh experiment (Sec.~\ref{exp:celea_bunny}), only adapt the dataset to handle the batch shape with different number of vertices. To be able to handle shape with large number of vertices, we pad all the shape and related Laplacian matrix to 60,000. During training, we set the batchsize to 4, and train on 4 GPUs for 96 epochs, which is approximately 48 hours.

\subsubsection{Qualitative results}
\textbf{Generative qualitative along the geometry.} Figure~\ref{fig:shapenet_tex} showcases qualitative results from our per-category texture generation experiment on the ShapeNetCore ``chair" category. Each column represents a different chair model, where DoubleDiffusion generates textures that adapt to the unique geometric structure of each mesh. The leftmost images in each row show the untextured chair models. Fig.~\ref{fig:shapenet_tex} demonstrate the qualitative result of texture generation on the different input chairs. The generated color on the mesh are smoothly varying following the chairs geometry, and the generated texture demonstrated a certain level of the understanding to the geometry. 

In each example, the generated color on the mesh smoothly varies in alignment with the geometry of the chair, demonstrating the model's capability to respect and adapt to surface details. Zoomed-in regions highlight the continuity and consistency of the generated textures, especially in curved and intricate parts of the mesh, such as the chair backs, armrests, and legs. More generation results on different category in is Fig~\ref{fig:multi-cate}, each category are trained independently.
\begin{figure}[t]
  \centering
    \includegraphics[width=\linewidth]{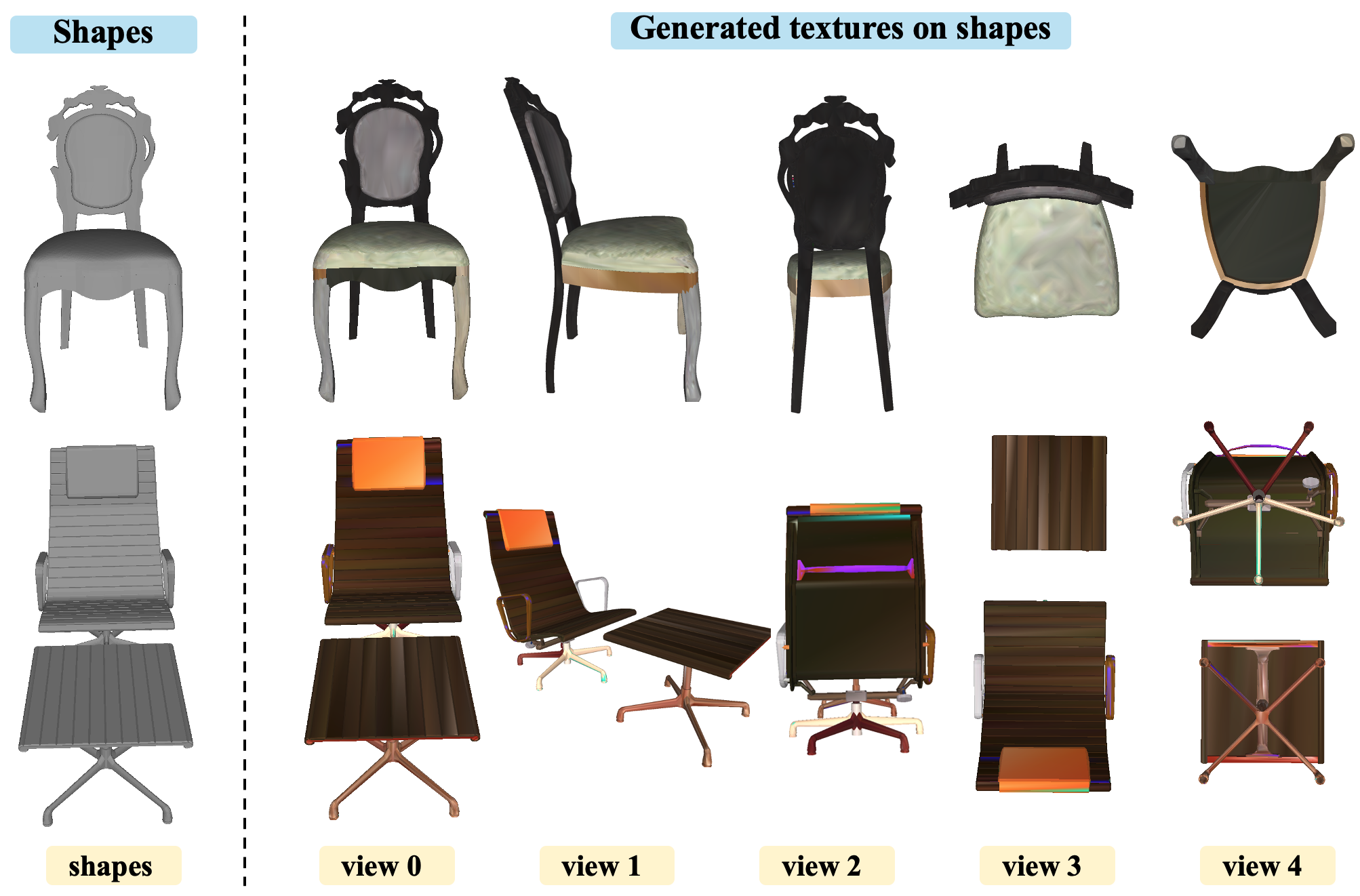}
    \caption{Generative result of the textured chairs from five distinct view points.} 
  \label{fig:shapenet_multiview}
\end{figure}
\textbf{View Consistency.} We present the generated results on chair models from five different views in Fig.~\ref{fig:shapenet_multiview}. This multi-view presentation demonstrates DoubleDiffusion's ability to achieve view consistency in signal generation directly on the mesh, without the need for 2D unrolling of the mesh. This consistency across views highlights the model’s capacity to respect the underlying mesh structure.

\section{Limitation and Future Work}
While the multi-view results demonstrate DoubleDiffusion’s strong capability for view-consistent texture generation directly on the mesh, it is also worth noting that our vertex-based approach relies heavily on vertex connectivity and the mesh Laplacian. This reliance can present challenges in accurately capturing color details on large faces, highlighting an area for future refinement in handling textures over varying face sizes on the mesh.  Meanwhile, our experiment can only demonstrate the per-category shape conditioned texture generation, lacking the ability to generalize to texture generation on multi-class shapes. The limitation comes from the lack of enough shape-texture data pair, which could potentially be solved by leveraging 2D methods for texture supervision. 

\section{Conclusion}
In this work, we introduced DoubleDiffusion, a novel framework that integrates heat dissipation diffusion with denoising diffusion models to directly generate geometry-respecting signal distributions on 3D meshes. Key contributions include a novel direct learning framework for diffusion on 3D manifold surfaces with the integration of heat dissipation diffusion for effective denoising aligned with mesh structure. We demonstrated that the propose framework performs geometry-adaptive signal generation across multiple complex surfaces.

{
    \small
    \bibliographystyle{ieeenat_fullname}
    \bibliography{main}
}


\end{document}